\documentclass[11pt,a4paper]{article}
\long\def\eat#1{}
\usepackage[hyperref]{acl2018}
\usepackage{times}
\usepackage{latexsym}

\usepackage{url}

\usepackage{extarrows}
\usepackage{amssymb,amsfonts}
\usepackage{latexsym}
\usepackage{multirow}
\usepackage{arydshln}
\usepackage{amsmath}
\usepackage{rotating}
\usepackage{color}
\usepackage{subfigure}

\aclfinalcopy 

\title{End-Task Oriented Textual Entailment \\ via Deep Explorations of
Inter-Sentence Interactions}

\author{Wenpeng Yin,$^1$ Hinrich Sch\"{u}tze,$^2$  Dan
  Roth$^{1}$ \\
  $^1$Dept.\ of Computer Science, University of
  Pennsylvania, USA\\
  $^2$CIS, LMU Munich, Germany\\
  {\tt
    \{wenpeng,danroth\}@seas.upenn.edu}}

\date{}

\newcommand{\dataname}{\textsc{Sci\-Tail}}
\newcommand{\modelname}{\textsc{DeIsTe}}

\newcounter{notecounter}

\newcommand{\enoteson}{\long\gdef\enote##1##2{{
      \stepcounter{notecounter}
                  {\large\bf
                    \hspace{1cm}\arabic{notecounter} $<<<$
                    ##1: ##2
                    $>>>$\hspace{1cm}}}}}
\enoteson

\def\mathlinebreak{\\[0.1cm]}
\def\mathindent{\mbox{\hspace{0.5cm}}}

\begin{document}
\maketitle
\begin{abstract}
This work deals with \dataname, a natural entailment
challenge derived from a multi-choice question answering
problem. The premises and hypotheses in \dataname\enspace
were
generated with no awareness of each other, and did not specifically aim at the
entailment task. This makes it
more challenging than other entailment data sets
and more
directly useful to the end-task -- question answering. We
propose \modelname\enspace (\textbf{d}eep 
\textbf{e}xplorations of
\textbf{i}nter-\textbf{s}entence interactions for
\textbf{t}extual \textbf{e}ntailment) for this entailment
task. Given word-to-word interactions between the
premise-hypothesis pair ($P$, $H$), \modelname\enspace
consists of: (i) a \emph{parameter-dynamic convolution} to
make important words in $P$ and $H$ play a dominant role in
learnt representations; and (ii) a \emph{position-aware
  attentive convolution} to encode the representation and
position information of the aligned word pairs.  Experiments
show that 
 \modelname\enspace gets $\approx$5\% improvement over
prior state of the
art and that the pretrained \modelname\enspace on \dataname\enspace generalizes well on RTE-5.\footnote{\url{https://github.com/yinwenpeng/SciTail}}
\end{abstract}


\section{Introduction}\label{sec:intro}
Textual entailment (TE) is a fundamental problem in natural language understanding and has been studied intensively recently using multiple benchmarks -- PASCAL RTE challenges
\cite{DBLPDaganGM05,DRSZ13}, Paragraph-Headline
\cite{Burger1631871}, SICK \cite{DBLPMarelliMBBBZ14} and
SNLI \cite{DBLPBowmanAPM15}.  In particular, SNLI -- while
much easier than earlier datasets -- has generated much work based on deep neural
networks due to its
large size. However, these benchmarks were mostly derived independently of any NLP problems.\footnote{RTE-\{5,6,7\} is an exception to this rule.} Therefore, the  premise-hypothesis pairs were composed under the constraint of
predefined rules and the language skills of humans. As a result,
while top-performing systems  push forward the
state-of-the-art, they do not necessarily  learn to support language inferences that  emerge commonly and naturally in  real NLP problems.

\eat{
\begin{table*}
 \setlength{\belowcaptionskip}{-15pt}
 \setlength{\abovecaptionskip}{5pt}
  \centering
  \begin{tabular}{c|l|c}
  Hypothesis $H$ &  \multicolumn{1}{c|}{ Premise $P$ } & label\\\hline
\multirow{4}{5cm}{Earth rotates on its axis one time in one day.} & Pluto rotates once on its axis every 6.39 Earth days. & 0\\
& Once per day, the earth rotates about its axis. & 1\\
&	It rotates on its axis once every 60 Earth days.&	0\\
& Earth orbits Sun, and rotates once per day about axis. & 1
\end{tabular}
\caption{Examples of \textsc{SciTail} dataset. Label ``1'' means \emph{entail}, ``0'' otherwise.}\label{tab:dataexample}
\end{table*}
}

In this work, we study \dataname\enspace \cite{scitail}, an
end-task oriented challenging entailment benchmark. \dataname\
is reformatted from a multi-choice question
answering problem.  All hypotheses $H$ were
obtained by rewriting  (question, correct answer) pairs;
all premises $P$ are relevant web sentences collected by an
Information retrieval (IR) method; then ($P$, $H$) pairs are
annotated via crowdsourcing. Table \ref{tab:dataexample}
shows examples. 
\emph{By this construction,  a substantial
  performance gain on \dataname\enspace can be  turned
  into better QA performance} \cite{scitail}.
\newcite{scitail} report that \dataname\enspace challenges
neural entailment models that show
outstanding performance on SNLI, e.g., Decomposable
Attention Model \cite{DBLParikhT0U16} and Enhanced LSTM
\cite{DBLPChenZLWJI17}.


\begin{table}
 \setlength{\belowcaptionskip}{-15pt}
 \setlength{\abovecaptionskip}{5pt}
  \centering
  \footnotesize
  \begin{tabular}{l|c}
   \multicolumn{1}{c|}{ Premise $P$ } & \\\hline
Pluto rotates once on its axis every 6.39 Earth days. & 0\\
 Once per day, the earth rotates about its axis. & 1\\
	It rotates on its axis once every 60 Earth days.&	0\\
 Earth orbits Sun, and rotates once per day about axis. & 1
\end{tabular}
  \caption{Examples of
    four premises for the hypothesis
    ``Earth rotates on its axis once times in one day'' in
    \textsc{SciTail} dataset. Right column (label): ``1'' means \emph{entail}, ``0'' otherwise.}\label{tab:dataexample}
\end{table}

\begin{figure}
 \setlength{\belowcaptionskip}{-12pt}
 \setlength{\abovecaptionskip}{5pt}
\centering
\includegraphics[width=0.4\textwidth]{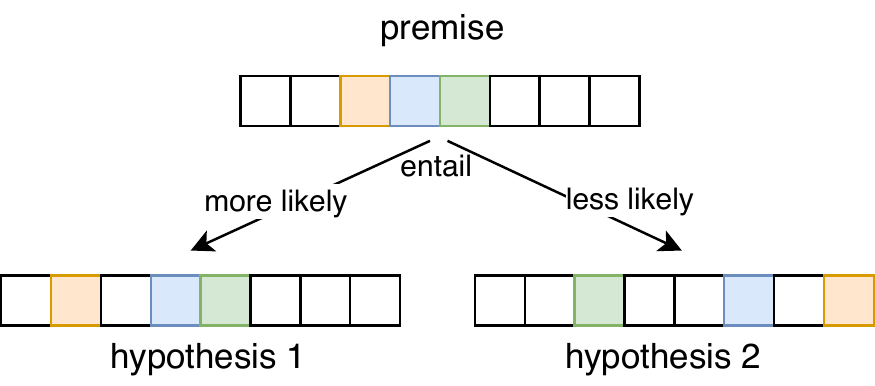}
\caption{The motivation of considering alignment positions in TE. The same color in (premise, hypothesis) means the two words are best aligned.}\label{fig:position}
\end{figure}

We propose \modelname\enspace for  \dataname.
Given word-to-word inter-sentence interactions between the pair ($P$,
$H$), \modelname\ pursues three deep exploration strategies of these
interactions. (a) How to express the
importance of a word, and let it play a dominant role in
learnt representations. (b) For any word in one of ($P$,
$H$), how to find the best aligned word in the other
sentence, so that we know their connection is indicative of
the final decision. (c) For a window of words in $P$ or $H$,
whether the locations of their best aligned words in the
other sentence provides clues. As Figure \ref{fig:position}
illustrates, the premise ``in this incident, the
cop ($C$) shot ($S$) the thief ($T$)'' is more likely to entail
the hypothesis ``$\hat{C} \ldots\ \hat{S} \ldots\ \hat{T}$'' than
``$\hat{T} \ldots\ \hat{S} \ldots\ \hat{C}$'' where $\hat{X}$ is the word
that best matches $X$.



Our model \modelname\enspace is implemented in convolutional
neural architecture \cite{lecun1998gradient}. Specifically,
\modelname\enspace consists of (i) a parameter-dynamic
convolution for exploration strategy (a) given above; and (ii) a
position-aware attentive convolution for exploration strategies
(b) and (c). In experiments, \modelname\enspace
outperforms prior top systems by $\approx$5\%. Perhaps even more interestingly, the pretrained model over \dataname\enspace generalizes well on RTE-5.

\begin{figure}
 \setlength{\belowcaptionskip}{-15pt}
 \setlength{\abovecaptionskip}{5pt}
\centering
\includegraphics[width=0.48\textwidth]{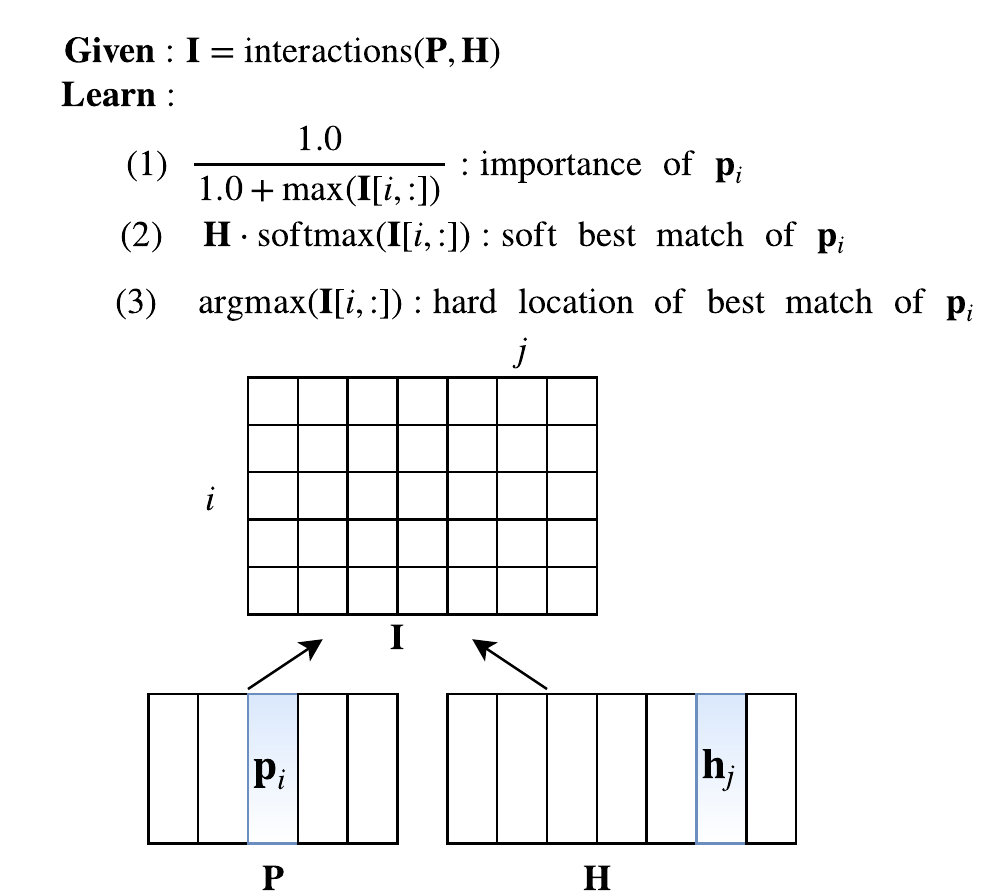}
\caption{The basic principles of \modelname\enspace in modeling the pair ($P$, $H$)}\label{fig:bigpic}
\end{figure}

\section{Method}


To start, a sentence $S$ ($S \in \{P,H\}$) is represented as a sequence of
$n_s$ hidden states, e.g.,  $\textbf{p}_i,\textbf{h}_i\in\mathbb{R}^d$
($i=1,2,\ldots, |n_{p/h}|$), forming a feature map
$\mathbf{S}\in\mathbb{R}^{d\times |n_s|}$, where $d$ is the
dimensionality of hidden states. Figure \ref{fig:bigpic} depicts the basic principle of  \modelname\enspace in modeling premise-hypothesis pair ($P$, $H$) with feature maps $\mathbf{P}$ and $\mathbf{H}$, respectively.

First, $\mathbf{P}$ and $\mathbf{H}$  have fine-grained interactions $\mathbf{I}\in\mathbb{R}^{n_p\times n_h}$ by comparing any pair of ($\mathbf{p}_i$,$\mathbf{h}_j$):
\begin{equation}
\abovedisplayskip=5pt
\belowdisplayskip=5pt
\mathbf{I}[i,j] = \mathrm{cosine}(\mathbf{p}_i, \mathbf{h}_j)
\end{equation}
We now elaborate \modelname's  exploration strategies (a), (b)
and (c) of the interaction results $\mathbf{I}$.

\begin{figure}
 \setlength{\belowcaptionskip}{-15pt}
 \setlength{\abovecaptionskip}{1pt}
\centering
\includegraphics[width=0.4\textwidth]{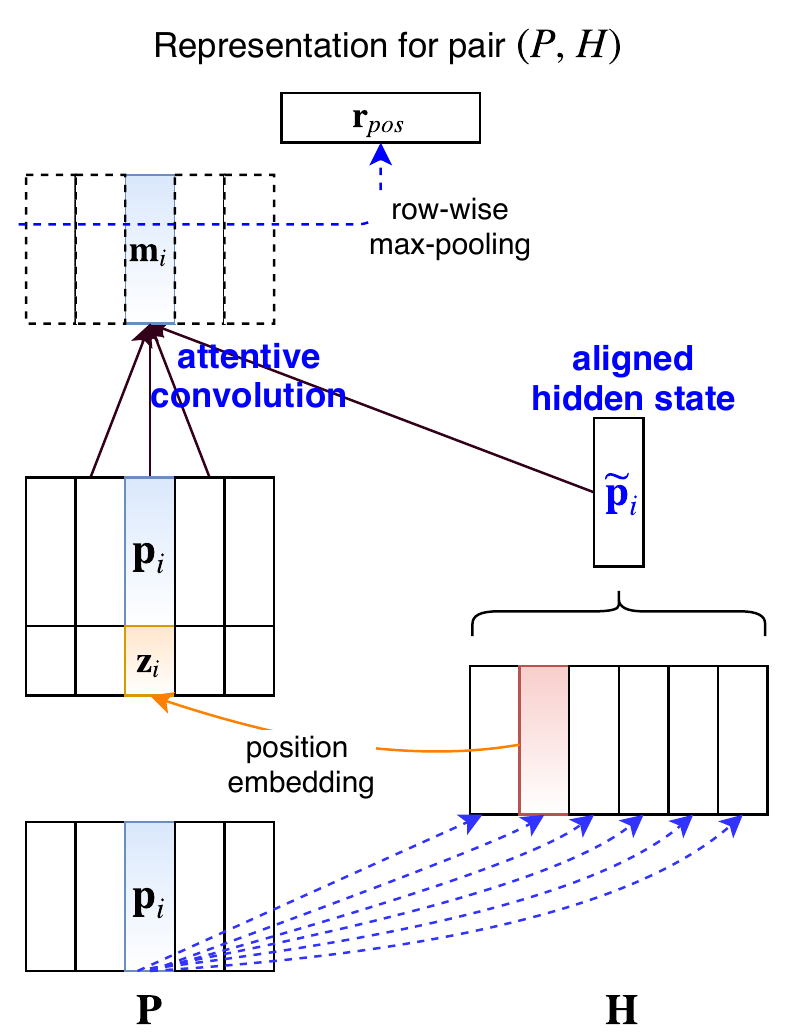}
\caption{Position-aware attentive convolution in modeling the pair ($P$, $H$)}\label{fig:lightacnn}
\end{figure}
\subsection{Parameter-dynamic convolution}
Intuitively, important words should be expressed more
intensively than other words in the learnt representation
of a sentence. However, the importance of words within a
specific sentence might not depend on the sentence
itself. E.g., \newcite{DBLPSchutzeY17a} found that in
question-aware answer sentence selection, words well matched
are more important; while in textual entailment, words
hardly matched are more important.

In this work, we try to make the semantics of those important words dominate in the output representations of a convolution encoder. 

Given a local window of hidden states in the feature map $\mathbf{P}$, e.g., three adjacent ones $\mathbf{p}_{i-1}$, $\mathbf{p}_i$ and $\mathbf{p}_{i+1}$, a conventional convolution   learns a higher-level representation $\mathbf{r}$ for this trigram:
\begin{equation}
\label{eq:conv}
\abovedisplayskip=5pt
\belowdisplayskip=5pt
\mathbf{r} = \mathrm{tanh}(\mathbf{W}\cdot[\mathbf{p}_{i-1}, \mathbf{p}_i,\mathbf{p}_{i+1}] + \mathbf{b})
\end{equation}
where $\mathbf{W}\in\mathbb{R}^{d\times 3d}$ and $\mathbf{b}\in\mathbb{R}^d$. 

For simplicity, we neglect the bias term $\mathbf{b}$ and split the multiplication of big matrices -- $\mathbf{W}\cdot[\mathbf{p}_{i-1}, \mathbf{p}_i,\mathbf{p}_{i+1}]$ -- into three parts, then $\mathbf{r}$ can be formulated as:
\begin{align*}
\mathbf{r}&=\mathrm{tanh}(\mathbf{W}^{-1}\cdot\mathbf{p}_{i-1}\oplus\mathbf{W}^{0}\cdot\mathbf{p}_i\oplus\mathbf{W}^{+1}\cdot\mathbf{p}_{i+1})\\
&=\mathrm{tanh}(\mathbf{\hat{p}}_{i-1}\oplus\mathbf{\hat{p}}_{i}\oplus\mathbf{\hat{p}}_{i+1})
\end{align*}
where $\oplus$ means element-wise addition; $\mathbf{W}^{-1}$, $\mathbf{W}^{0}$, $\mathbf{W}^{+1}\in\mathbf{R}^{d\times d}$, and their concatenation equals to the $\mathbf{W}$ in Equation \ref{eq:conv}; $\mathbf{\hat{p}}_{i}$ is set to be $\mathbf{W}^{0}\cdot\mathbf{p}_i$, so $\mathbf{\hat{p}}_{i}$ can be seen as the projection of $\mathbf{p}_i$ in a new space by parameters $\mathbf{W}^{0}$; finally the three projected outputs contribute equally in the addition: $\mathbf{\hat{p}}_{i-1}\oplus\mathbf{\hat{p}}_{i}\oplus\mathbf{\hat{p}}_{i+1}$.

The convolution encoder shares parameters
[$\mathbf{W}^{-1}$, $\mathbf{W}^{0}$, $\mathbf{W}^{+1}$] in
all trigrams, so we cannot expect those parameters to
express the importance of $\mathbf{\hat{p}}_{i-1}$,
$\mathbf{\hat{p}}_{i}$ or $\mathbf{\hat{p}}_{i+1}$ in the
output representation $\mathbf{r}$. Instead, we  formulate
the idea as follows:\mathlinebreak
\mathindent $\mathbf{m}_{i} = \mathrm{tanh}(\alpha_{i-1}\mathbf{\hat{p}}_{i-1}\oplus\alpha_{i}\mathbf{\hat{p}}_{i}\oplus\alpha_{i+1}\mathbf{\hat{p}}_{i+1})$\mathlinebreak
in which the three scalars $\alpha_{i-1}$, $\alpha_{i}$ and $\alpha_{i+1}$ indicate the importance scores of $\mathbf{\hat{p}}_{i-1}$, $\mathbf{\hat{p}}_{i}$ and $\mathbf{\hat{p}}_{i+1}$ respectively. In our work, we adopt:
\begin{equation}
\alpha_{i} = \frac{1.0}{1.0+\mathrm{max}(\mathbf{I}[i,:])}
\end{equation}

Since
$\alpha_{i}\mathbf{\hat{p}}_{i}=\alpha_{i}\mathbf{W}^{0}\cdot\mathbf{p}_i
= \mathbf{W}^{0,i}\cdot\mathbf{p}_i$, we notice that the
original shared parameter $\mathbf{W}^{0}$ is mapped to a
dynamic parameter $\mathbf{W}^{0,i}$, which is specific to
the input $\mathbf{p}_{i}$. We refer to this as 
\emph{parameter-dynamic convolution}, which enables our
system \modelname\enspace to highlight important words in
higher-level representations.

Finally, a max-pooling layer is stacked over \{$\mathbf{m}_i$\} to get the representation for the pair ($P$, $H$), denoted as $\mathbf{r}_{dyn}$.


\begin{table}
 \setlength{\belowcaptionskip}{-8pt}
 \setlength{\abovecaptionskip}{5pt}
  \centering
  \begin{tabular}{l|cc}
  methods & dev & test\\\hline
  Majority Class &	50.4	&60.4\\
  Hypothesis only & 66.9& 65.1\\
  Premise only & 72.6& 73.4\\\hdashline

  NGram model  &	65.0&	70.6\\
  ESIM-600D 	&70.5&	70.6\\
  Decomp-Att  &	75.4	&72.3\\
  DGEM &	79.6	&77.3\\
  AttentiveConvNet  &79.3& 78.1\\\hline
  \modelname &\textbf{82.4} & \textbf{82.1}\\
  \enspace\enspace w/o dyn-conv & 80.2& 79.1\\
  \enspace\enspace w/o representation & 76.3& 74.9\\
    \enspace\enspace w/o position & 82.1& 81.3\\
\end{tabular}
\caption{\modelname\enspace vs.\  baselines on  \dataname}\label{tab:results}
\end{table}
\subsection{Position-aware attentive convolution}
Our position-aware attentive convolution, shown in Figure  \ref{fig:lightacnn}, aims to encode the representations as well as the positions of the best word alignments.

\textbf{Representation.} 
Given the interaction scores in $\mathbf{I}$, the representation $\tilde{\mathbf{p}}_i$ of all soft matches   for hidden state $\mathbf{p}_i$ in $P$ is the weighted average of all hidden states in $H$:
\begin{equation}
\abovedisplayskip=2pt
\belowdisplayskip=2pt
\tilde{\mathbf{p}}_i = \sum_j \mathrm{softmax}(\mathbf{I}[i,:])_j\cdot \mathbf{h}_j
\end{equation}

\textbf{Position.}
 For a trigram [$\mathbf{p}_{i-1}$, $\mathbf{p}_{i}$,
   $\mathbf{p}_{i+1}$] in $P$, knowing where its
 best-matched words are located in $H$ is helpful in TE, as  discussed in Section \ref{sec:intro}.

For $\mathbf{p}_{i}$, we first retrieve the index $x_i$ of the best-matched word in $H$ by:
\begin{equation}
x_i = \mathrm{argmax}(\mathbf{I}[i,:])
\end{equation}
then embed the concrete \{$x_i$\} into randomly-initialized continuous space:
\begin{equation}
\mathbf{z}_i = \mathbf{M}[x_i]
\end{equation}
where  $\mathbf{M}\in\mathbb{R}^{n_h\times d_m}$; $n_h$ is the length of $H$; $d_m$ is the dimensionality of position embeddings.

Now, the three positions [$i-1$, $i$, $i+1$] in $P$ concatenate vector-wisely original hidden states [$\mathbf{p}_{i-1}$, $\mathbf{p}_{i}$, $\mathbf{p}_{i+1}$] with position embeddings [$\mathbf{z}_{i-1}$, $\mathbf{z}_{i}$, $\mathbf{z}_{i+1}$], getting a new sequence of hidden states: [$\mathbf{c}_{i-1}$, $\mathbf{c}_{i}$, $\mathbf{c}_{i+1}$]. As a result, a position $i$ in $P$ has hidden state $\mathbf{c}_i$,  left
context $\mathbf{c}_{i-1}$, right context
$\mathbf{c}_{i+1}$ and the representation of soft-aligned words in $H$, i.e., 
$\tilde{\mathbf{p}}_i$. Then a convolution works at position $i$ in $P$ as:
\begin{align}\label{eq:rpos}
\abovedisplayskip=0pt
\belowdisplayskip=0pt
\mathbf{n}_i=\mathrm{tanh}(&\mathbf{W}\cdot [\mathbf{c}_{i-1},\mathbf{c}_i,\mathbf{c}_{i+1},\tilde{\mathbf{p}}_i]+\mathbf{b}) 
\end{align}

As Figure \ref{fig:lightacnn} shows, the position-aware attentive convolution finally stacks a standard max-pooling layer over \{$\mathbf{n}_i$\} to get the representation for the pair ($P$, $H$), denoted as $\mathbf{r}_{pos}$. 

Overall, our \modelname\enspace will generate a representation $\mathbf{r}_{dyn}$ through the parameter-dynamic convolution, and generate a representation $\mathbf{r}_{pos}$ through the position-aware attentive convolution. Finally the concatenation --  [$\mathbf{r}_{dyn}$, $\mathbf{r}_{pos}$] -- is fed to a logistic regression classifier.

\section{Experiments}

\textbf{Dataset.} \textsc{SciTail}\footnote{Please refer to \cite{scitail} for more  details.}  \cite{scitail} is for textual entailment in binary classification: entailment or neutral. Accuracy is reported.

\textbf{Training setup.} 
All words are initialized by 300D  Word2Vec \cite{mikolov2013distributed} embeddings, and are fine-tuned during training. The whole system is trained by AdaGrad \cite{duchi2011adaptive}. Other hyperparameter values include: learning rate 0.01, $d_m$=50 for  position embeddings $\mathbf{M}$, hidden size 300, batch size 50, filter width 3. 

\begin{table}
 \setlength{\belowcaptionskip}{-15pt}
 \setlength{\abovecaptionskip}{5pt}
  \centering
  \begin{tabular}{l|c}
  methods & acc.\\\hline
  Majority Class &	34.3\\
Premise only & 33.5\\
Hypothesis only & 68.7\\\hdashline
  ESIM 	&86.7\\
  Decomp-Att  &	86.8\\
  AttentiveConvNet  &86.3\\\hline
  \modelname$_{\dataname}$ & 84.7\\
    \modelname$_{tune}$ & 87.1\\\hline
    State-of-the-art & 88.7
\end{tabular}
\caption{\modelname\enspace vs.\  baselines on  SNLI.  \modelname$_{\dataname}$  has exactly the same system layout and hyperparameters as the one reported on \dataname\enspace in Table \ref{tab:results}; \modelname$_{tune}$: tune hyperparameters on SNLI dev. State-of-the-art refers to \cite{DBLP05365}. Ensemble results are not considered.}\label{tab:snliresults}
\end{table}
\begin{table*}
\setlength{\tabcolsep}{3pt}
 \setlength{\belowcaptionskip}{-10pt}
 \setlength{\abovecaptionskip}{5pt}
  \centering
  \small
  \begin{tabular}{c|l|c|l}
   \# & \multicolumn{1}{c|}{(Premise $P$, Hypothesis $H$) Pair} & G/P & Challenge \\\hline
  
   \multirow{2}{*}{1} & ($P$) Front -- The boundary between two different air masses. & \multirow{2}{*}{1/0} & \multirow{2}{1.5cm}{language conventions}\\
  & ($H$) In weather terms, the boundary between two air masses is called front. & & \\\hline
  
    \multirow{2}{*}{2} & ($P$) \ldots\ the notochord forms the backbone (or vertebral column). & \multirow{2}{*}{1/0} & \multirow{2}{1.5cm}{language conventions}\\
  & ($H$) Backbone is another name for the vertebral column. & & \\\hline
  
        \multirow{2}{*}{3} & ($P$) $\cdots$  animals with a vertebral column or backbone and animals without one. & \multirow{2}{*}{1/0} & \multirow{2}{*}{ambiguity}\\
  & ($H$) Backbone is another name for the vertebral column. & &\\\hline
  
  
      \multirow{3}{*}{4} & \multirow{2}{10.5cm}{($P$) Heterotrophs get energy and carbon from living plants or animals ( consumers ) or from dead organic matter ( decomposers ).} & \multirow{3}{*}{0/1} & \multirow{3}{1.5cm}{discourse relation}\\
  &  & & \\
  & ($H$) Mushrooms get their energy from decomposing dead organisms. & &\\\hline

      \multirow{3}{*}{5} & \multirow{2}{10.5cm}{($P$) Ethane is a simple hydrocarbon, a molecule made of two carbon and six hydrogen atoms.} & \multirow{3}{*}{0/1} & \multirow{3}{1.5cm}{discourse relation}\\
  &  & & \\
  & ($H$) Hydrocarbons are made of one carbon and four hydrogen atoms. & &\\\hline
  
      \multirow{2}{*}{6} & ($P$) \ldots\ the SI unit\ldots for force is the Newton (N) and is defined as (kg$\cdot$m/s$^{-2}$ ). & \multirow{2}{*}{0/1} & \multirow{2}{*}{beyond text}\\
  & ($H$) Newton (N) is the SI unit for weight. & &

\end{tabular}
\caption{Error cases of \modelname\enspace in \dataname. ``$\cdots$'': truncated text. ``G/P'': gold/predicted label.}\label{tab:errorexample}
\end{table*}

\textbf{Baselines.}
(i) Decomposable Attention Model (Decomp-Att)
\cite{DBLParikhT0U16}: Develop attention mechanisms to
decompose the problem into  subproblems to solve in
parallel. (ii) Enhanced LSTM (ESIM) \cite{DBLPChenZLWJI17}:
Enhance LSTM by encoding syntax and semantics from parsing
information. (iii) Ngram Overlap: An overlap baseline,
considering unigrams, one-skip bigrams and one-skip
trigrams.  (iv) DGEM \cite{scitail}: A decomposed graph
entailment model, the current state-of-the-art. (v) AttentiveConvNet \cite{DBLP00519}: Our top-performing textual entailment system on SNLI dataset, equipped with RNN-style attention mechanism in convolution.\footnote{https://github.com/yinwenpeng/Attentive\_Convolution}

In addition, to check if \dataname\enspace can be easily
resolved by features from only premises or hypotheses (like
the problem of SNLI shown by \newcite{DBLP02324}), we put a
vanilla CNN (convolution\&max-pooling) over merely
hypothesis or premise to derive the pair label.

Table \ref{tab:results} presents \textbf{results} on \dataname.
 (i) Our model \modelname\enspace has a big improvement
($\sim$ 5\%) over DGEM, the best baseline.
Interestingly, AttentiveConvNet performs very
competitively, surpassing DGEM by 0.8\% on test. These two results show
the effectiveness of attentive convolution. \modelname,
equipped with a parameter-dynamic convolution and a more
advanced position-aware attentive convolution, clearly gets
a big plus. (ii) The ablation  shows that all three aspects
we explore from the inter-sentence interactions
contribute; ``position'' encoding is less important than
``dyn-conv'' and ``representation''. Without
``representation'', the system performs much worse. This
is in line with the result of AttentiveConvNet baseline.

To further study the systems and datasets,
Table \ref{tab:snliresults} gives
performance of \modelname\enspace and baselines on SNLI.
We see that  \modelname\enspace  gets
competitive performance on SNLI.

Comparing Tables \ref{tab:results} and \ref{tab:snliresults},
the baselines ``hypothesis only'' and ``premise only'' show
analogous while different phenomena between
\dataname\enspace and SNLI. On one hand, both SNLI and
\dataname\enspace can get a relatively high number by
looking at only one of \{premise, hypothesis\} -- ``premise
only'' gets 73.4\% accuracy on \dataname,  even higher than
two more complicated baselines (ESIM-600D and Decomp-Att),
and ``hypothesis only'' gets 68.7\% accuracy on SNLI which
is more than 30\% higher than the ``majority''  and
``premise only'' baselines. Notice the contrast: SNLI ``prefers'' hypothesis, \dataname\enspace ``prefers'' premise. For SNLI, this is not surprising as the crowd-workers tend to construct the  hypotheses in SNLI by some regular rules \cite{DBLP02324}. The phenomenon in \dataname\enspace is left to explore in future work.

\textbf{Error Analysis.}  Table \ref{tab:errorexample}
gives examples of errors. We explain them as follows.

\emph{Language conventions}: The pair \#1 uses dash ``--'' to indicate a definition sentence for ``Front''; The pair \#2 has ``A (or B)'' to denote the equivalence between A and B. This challenge is expected to be handled by rules.

\emph{Ambiguity}: The pair \#3 looks like having a similar
challenge with the pair \#2. We guess the annotators treat
``$\cdots$ a vertebral column or backbone'' and `` $\cdots$
the backbone (or vertebral column)'' as the same convention, which may be debatable.


\emph{Complex discourse relation}: The premise in the pair \#4 has an ``or'' structure. In the pair \#5, ``a molecule made of $\cdots$'' defines the concept ``Ethane'' instead of the ``hydrocarbon''. Both cases require the model to be able to comprehend the discourse relation.

\emph{Knowledge beyond text}: The main challenge in the pair \#6 is to distinguish between ``weight'' and ``force'', which requires more physical knowledge that is beyond the text described here and beyond the expressivity of word embeddings.

\textbf{Transfer to RTE-5.} One main motivation of exploring
this \dataname\enspace problem is that this is an end-task
oriented TE task. A natural question thus is how well the
trained model can be transferred to other end-task oriented
TE tasks.  In Table \ref{tab:rte5results}, we take the
models pretrained on \dataname\enspace and SNLI and test
them on 
RTE-5. Clearly, the model pretrained on SNLI has not
learned anything useful for
RTE-5 -- its performance of 46.0\% is even worse than
the majority baseline. The model pretrained on \dataname,
in contrast, demonstrates much more promising
generalization performance: 60.2\% vs.\ 46.0\%.

\begin{table}
 \setlength{\belowcaptionskip}{-15pt}
 \setlength{\abovecaptionskip}{5pt}
  \centering
  \begin{tabular}{l|c c}
  & dev & test\\\hline
	Majority baseline & 50.0 & 50.0\\
    State-of-the-art  &-- & 73.5 \\\hline
    training data & &\\\hline
    SNLI & 47.1 & 46.0 \\
    \dataname & 60.5 & 60.2\\
    
\end{tabular}
\caption{Train on different TE datasets, test accuracy on two-way RTE-5. State-of-the-art refers to \cite{DBLPIfteneM09}}\label{tab:rte5results}
\end{table}

\section{Related Work}
Learning automatically inter-sentence word-to-word interactions or alignments was first studied in recurrent neural networks \cite{journalsElman90}.
 \newcite{entail2016}  employ neural word-to-word attention for SNLI task.  \newcite{wang2015learning} propose match-LSTM, an extension of the attention mechanism in \cite{entail2016}, by more fine-grained matching and accumulation. \newcite{DBLP0001DL16} present a new LSTM equipped with a memory tape. Other work following this attentive matching idea includes Bilateral Multi-Perspective Matching  model \cite{DBLPWangHF17}, Enhanced LSTM  \cite{DBLPChenZLWJ16} etc.

In addition, convolutional neural networks \cite{lecun1998gradient}, equipped with attention mechanisms, also perform competitively in TE. \newcite{DBLPYinSXZ16} implement the attention \emph{in pooling phase} so that important hidden states will be pooled with higher probabilities. \newcite{DBLP00519} further develop the attention idea in CNNs, so that a RNN-style attention mechanism is integrated into the convolution filters. This is similar with our position-aware attentive convolution. We instead explored a way to make use of position information of alignments to do reasoning.

Attention mechanisms in both RNNs and CNNs make use of sentence interactions. Our work achieves a deep exploration of those interactions, in order to guide representation learning in TE.

\section{Summary}
This work proposed \modelname\enspace to deal with an end-task oriented textual entailment task -- \dataname. Our model enables a comprehensive utilization of inter-sentence interactions. \modelname\enspace outperforms competitive systems by big margins.

\textbf{Acknowledgments.}
We thank all the reviewers for insightful comments. This research is supported in part by DARPA under agreement number FA8750-13-2-0008, and by a gift from Google. 

\bibliography{acl2018,ccg-compact}
\bibliographystyle{acl_natbib}
\end{document}